\documentclass[letterpaper]{article}
\usepackage{aaai19}
\usepackage{times}
\usepackage{helvet}
\usepackage{courier}
\usepackage{url}
\usepackage{graphicx}
\usepackage[utf8]{inputenc} 

\newif\ifblind
\newcommand{\blind}[1]{#1}

\title{Learning Vision-based Cohesive Flight in Drone Swarms}
\author{%
\blind{\nameschilling \tno{,} \namelecoeur \tno{,} \nameschiano \tno{, and} \namefloreano}\\
\blind{Laboratory of Intelligent Systems}\\
\blind{École Polytechnique Fédérale de Lausanne}\\
\blind{CH-1015 Lausanne, Switzerland}\\
\blind{\ttt{\{fabian.schilling,julien.lecoeur,fabrizio.schiano,dario.floreano\}@epfl.ch}}
}

\ifblind
\else
\nocopyright
\fi

\usepackage[english]{babel}
\usepackage{amsmath, amssymb, amsfonts}
\usepackage{booktabs}
\usepackage{color}
\usepackage{graphicx}
\usepackage{siunitx}
\usepackage{subcaption}

\graphicspath{{./figures/}}

\newcommand{\nameschilling}{Fabian Schilling}
\newcommand{\namelecoeur}{Julien Lecoeur}
\newcommand{\nameschiano}{Fabrizio Schiano}
\newcommand{\namefloreano}{Dario Floreano}

\newcommand{\tit}[1]{\textit{#1}}

\newcommand{\ttt}[1]{\texttt{#1}}
\newcommand{\tno}[1]{\textnormal{#1}}

\newcommand{\mbf}[1]{\mathbf{#1}}
\newcommand{\mcl}[1]{\mathcal{#1}}
\newcommand{\mbb}[1]{\mathbb{#1}}
\newcommand{\mtx}[1]{\text{#1}}
\newcommand{\txt}[1]{\text{#1}}

\newcommand{\Vector}[1]{\mbf{#1}}
\newcommand{\vel}{\Vector{v}}
\newcommand{\pos}{\Vector{p}}
\newcommand{\rel}{\Vector{r}}

\newcommand{\World}{\mcl{W}}
\newcommand{\Body}{\mcl{B}}

\newcommand{\SO}[1]{\txt{SO}(#1)}

\newcommand{\Rotation}{\mbf{R}}

\newcommand{\orderparam}{\omega^\txt{ord}}

\newcommand{\norm}[1]{\|#1\|}

\newcommand{\set}[1]{\mcl{#1}}
\newcommand{\agents}{\set{A}}
\newcommand{\neighbors}{\set{N}}

\newcommand{\Real}{\mbb{R}}

\newcommand{\lr}{\eta}
\newcommand{\wdecay}{\lambda}
\newcommand{\loss}{\mcl{L}}
\newcommand{\momentum}{\mu}
\newcommand{\pdropout}{p}
\newcommand{\batchsize}{\mcl{B}}
\newcommand{\pred}[1]{\hat{#1}}
\newcommand{\target}[1]{#1}

\begin{document}

\maketitle


\begin{abstract}
This paper presents a data-driven approach to learning vision-based collective behavior from a simple flocking algorithm.
We simulate a swarm of quadrotor drones and formulate the controller as a regression problem in which we generate 3D velocity commands directly from raw camera images.
The dataset is created by simultaneously acquiring omnidirectional images and computing the corresponding control command from the flocking algorithm.
We show that a convolutional neural network trained on the visual inputs of the drone can learn not only robust collision avoidance but also coherence of the flock in a sample-efficient manner.
The neural controller effectively learns to localize other agents in the visual input, which we show by visualizing the regions with the most influence on the motion of an agent.
This weakly supervised saliency map can be computed efficiently and may be used as a prior for subsequent detection and relative localization of other agents.
We remove the dependence on sharing positions among flock members by taking only local visual information into account for control.
Our work can therefore be seen as the first step towards a fully decentralized, vision-based flock without the need for communication or visual markers to aid detection of other agents.
\end{abstract}


\section{Introduction}\label{sec:introduction}

Collective motion of animal groups such as flocks of birds is an awe-inspiring natural phenomenon that has profound implications for the field of aerial swarm robotics \cite{floreano_science_2015,zufferey_aerial_2011}.
Animal groups in nature operate in a completely self-organized manner since the interactions between them are purely local and decisions are made by the animals themselves.
By taking inspiration from decentralization in biological systems, we can develop powerful robotic swarms that are 1) robust to failure, and 2) highly scalable since the number of agents can be increased or decreased depending on the workload.

One of the most appealing characteristics of collective animal behavior for robotics is that decisions are made based on local information such as visual perception.
As of today, however, most multi-agent robotic systems rely on entirely centralized control \cite{mellinger_minimum_2011,kushleyev_towards_2013,preiss_crazyswarm_2017,weinstein_visual_2018} or wireless communication of positions \cite{vasarhelyi_outdoor_2014,viragh_flocking_2014,vasarhelyi_optimized_2018}, either from a motion capture system or global navigation satellite system (GNSS).
The main drawback of these approaches is the introduction of a single point of failure, as well as the use of unreliable data links, respectively.
Relying on centralized control bears a significant risk since the agents lack the autonomy to make their own decisions in failure cases such as a communication outage.
The possibility of failure is even higher in dense urban environments, where GNSS measurements are often unreliable and imprecise.

\begin{figure}[t!]
    \centering
    \includegraphics[width=\columnwidth]{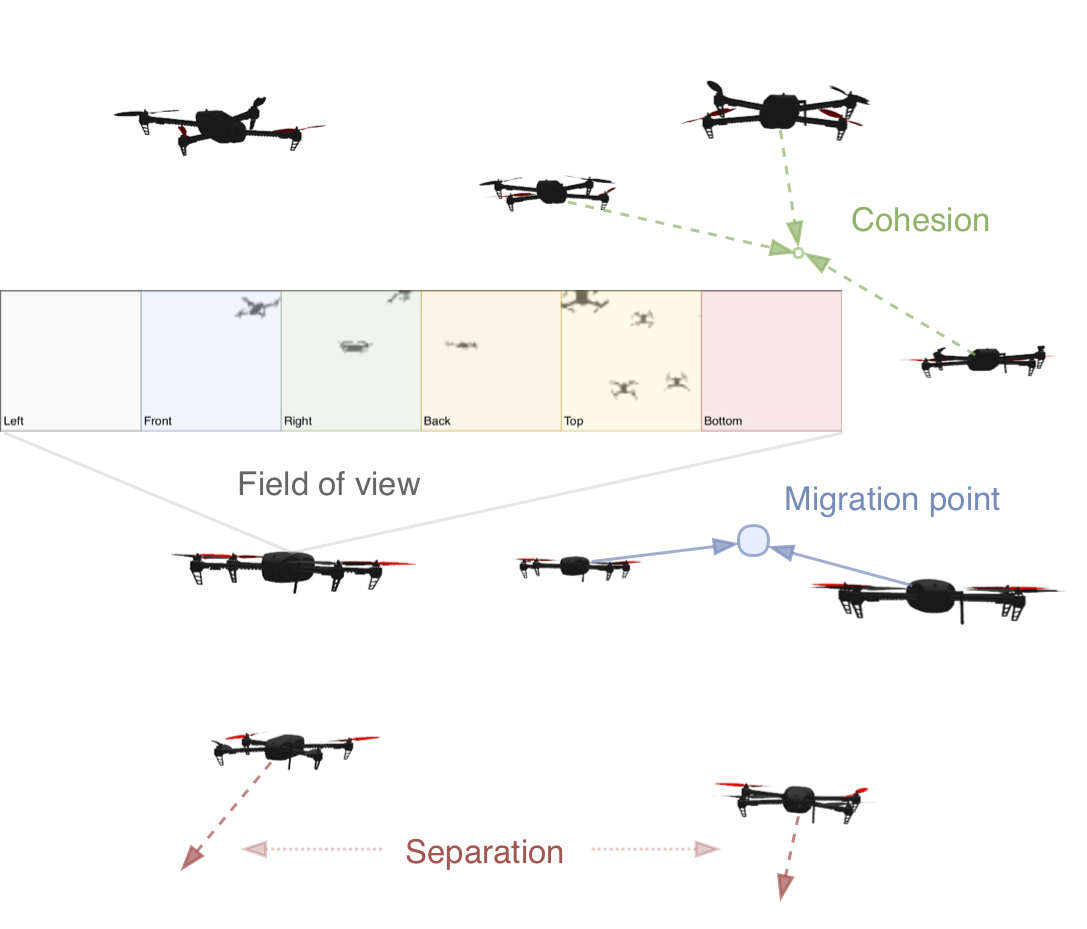}
    \caption{%
        Vision-based flock of nine drones during migration.
        Our visual swarm controller operates fully decentralized and provides collision-free, coherent collective motion without the need to share positions among agents.
        The behavior of an agent depends only on the omnidirectional visual inputs (colored rectangle, see Fig.~\ref{fig:camera-config} for details) and the migration point (blue circle and arrows).
        Collision avoidance (red arrows) and coherence (green arrows) between flock members are learned entirely from visual inputs.
    }\label{fig:overview}
\end{figure}

Vision is arguably the most promising sensory modality to achieve a maximum level of autonomy for robotic systems, particularly considering the recent advances in computer vision and deep learning \cite{krizhevsky_imagenet_2012,lecun_deep_2015,he_deep_2016}.
Apart from being light-weight and having relatively low power consumption, even cheap commodity cameras provide an unparalleled information density with respect to sensors of similar cost.
Their characteristics are specifically desirable for the deployment of an aerial multi-robot system.
The difficulty when using cameras for robot control is the interpretation of the visual information which is a hard problem that this paper addresses directly.

In this work, we propose a reactive control strategy based \textit{only} on local visual information.
We formulate the swarm interactions as a regression problem in which we predict control commands as a nonlinear function of the visual input of a single agent.
To the best of our knowledge, this is the first successful attempt to learn vision-based swarm behaviors such as collision-free navigation in an end-to-end manner directly from raw images.

\begin{figure*}[t]
    \centering
    \begin{subfigure}[b]{0.31\textwidth}
        \centering
        \includegraphics[width=\textwidth]{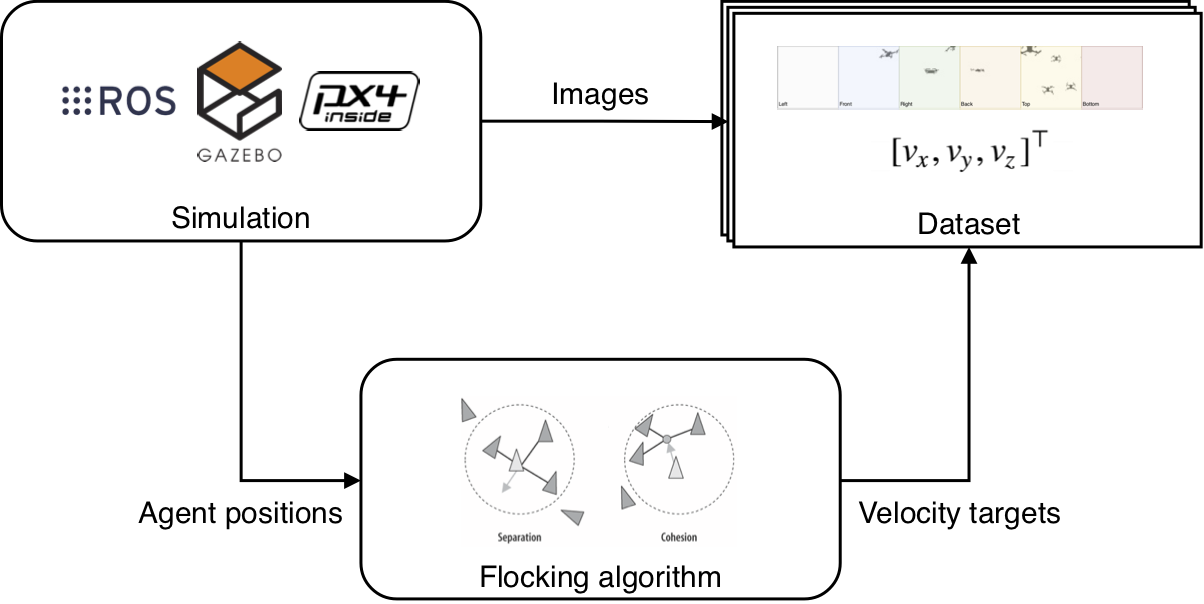}
        \caption{Dataset generation}\label{fig:dataset-generation}
    \end{subfigure}
    \hfill
    \begin{subfigure}[b]{0.31\textwidth}
        \centering
        \includegraphics[width=\textwidth]{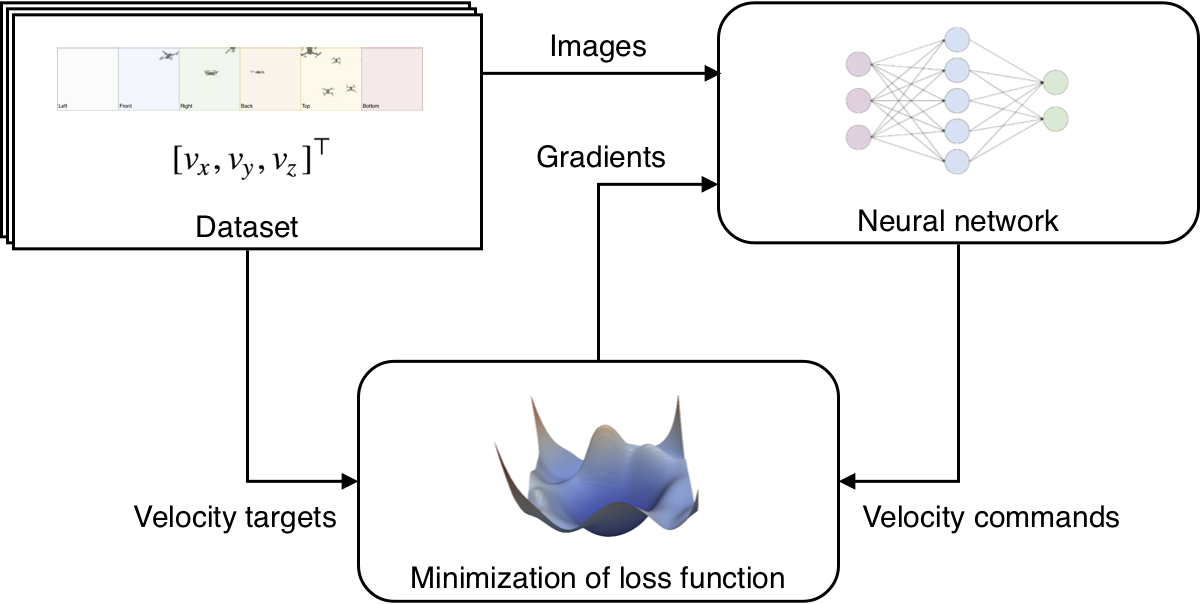}
        \caption{Training phase}\label{fig:training-phase}
    \end{subfigure}
    \hfill
    \begin{subfigure}[b]{0.31\textwidth}
        \centering
        \includegraphics[width=\textwidth]{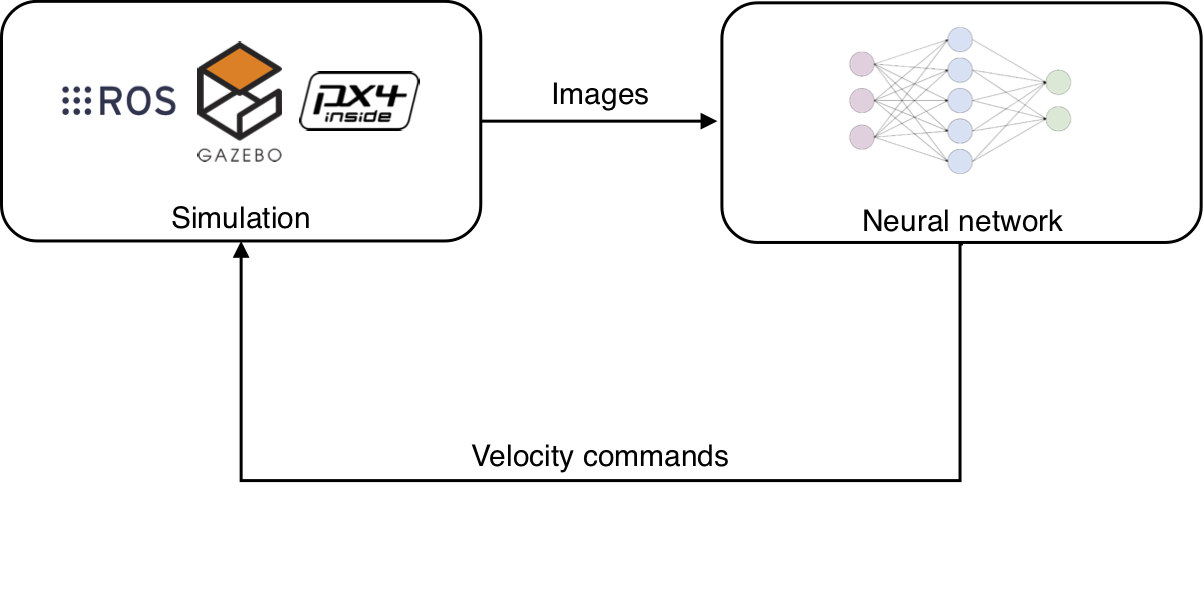}
        \caption{Vision-based control}\label{fig:visual-flocking}
    \end{subfigure}
    \caption{%
        Overview of our method.
        (\ref{fig:dataset-generation}) The dataset is generated using a simple flocking algorithm (see Sec.~\ref{sec:flocking-algorithm}) deployed on simulated quadrotor drones (see Sec.~\ref{sec:drone-model}).
        The agents are initialized in random non-overlapping positions and assigned random linear trajectories while computing velocity targets and acquiring images from all mounted cameras (see Sec.~\ref{sec:dataset-generation}).
        (\ref{fig:training-phase}) During the training phase (see Sec.~\ref{sec:training-phase}), we employ a convolutional neural network to learn the mapping between visual inputs and control outputs by minimizing the error between predicted and target velocity commands.
        (\ref{fig:visual-flocking}) During the last phase (see Sec.~\ref{sec:vision-based-control}), we use the trained network as a reactive controller for emulating vision-based swarm behavior.
        The behavior of each agent is entirely vision-based and only depends on its visual input at a specific time step.
        Initial flight experiments have confirmed that by using PX4 and ROS, we can use the exact same software infrastructure for both simulation and reality.
        }\label{fig:flowchart}
\end{figure*}


\section{Related Work}\label{sec:related-work}

We classify the related work into three main categories.
Sec.~\ref{sec:decentralized-flocking} considers literature in which a flock of drones is controlled in a fully decentralized manner.
Sec.~\ref{sec:vision-based} is comprised of recent data-driven advances in vision-based drone control.
Finally, Sec.~\ref{sec:vision-flocking} combines ideas from the previous sections into approaches that are both vision-based and decentralized.

\subsection{Decentralized flocking with drones}\label{sec:decentralized-flocking}

Flocks of autonomous drones such as quadrotors and fixed-wings are the focus of recent research in swarm robotics.
Early work presents ten fixed-wing drones deployed in an outdoor environment \cite{hauert_reynolds_2011}.
Their collective motion is based on Reynolds flocking \cite{reynolds_flocks_1987} with a migration term that allows the flock to navigate towards the desired goal.
Arising from the use of a nonholonomic platform, the authors study the interplay of the communication range and the maximum turning rate of the agents.

Thus far, the largest decentralized quadrotor flock consisted of 30 autonomous agents flying in an outdoor environment \cite{vasarhelyi_optimized_2018}.
The underlying algorithm has many free parameters which require the use of optimization methods.
To this end, the authors employ an evolutionary algorithm to find the best flocking parameters according to a fitness function that relies on several order parameters.
The swarm can operate in pre-defined confined space by incorporating repulsive virtual agents.

The commonality of all mentioned approaches and others, for example, \cite{vasarhelyi_outdoor_2014,dousse_human-comfortable_2017}, is the ability to share GNSS positions wirelessly among flock members.
However, there are many situations in which wireless communication is unreliable or GNSS positions are too imprecise.
We may not be able to tolerate position imprecisions in situations where the environment requires a small inter-agent distance, for example when traversing narrow passages in urban environments.
In these situations, tall buildings may deflect the signal and communication outages occur due to the wireless bands being over-utilized.

\subsection{Vision-based single drone control}\label{sec:vision-based}

Vision-based control of a single flying robot is facilitated by several recent advances in the field of machine learning.
In particular, the controllers are based on three types of learning methods: imitation learning, supervised learning, and reinforcement learning.

Imitation learning is used in \cite{ross_learning_2013} to control a drone in a forest environment based on human pilot demonstrations.
The authors motivate the importance of following suboptimal control policies in order to cover more of the state space.
The reactive controller can avoid trees by adapting the heading of the drone; the limiting factor is ultimately the field of view of a single front-facing camera.

A supervised learning approach \cite{loquercio_dronet_2018} features a convolutional network that is used to predict a steering angle and a collision probability for drone navigation in urban environments.
Steering angle prediction is formulated as a regression problem by minimizing the mean-squared-error between predicted and ground truth annotated examples from a dataset geared for autonomous driving research.
The probability of collision is learned by minimizing the binary cross-entropy of labeled images collected while riding a bicycle through urban environments.
The drone is controlled directly by the steering angle, whereas its forward velocity is modulated by the collision probability.

An approach based on reinforcement learning \cite{sadeghi_cad2rl_2017} shows that a neural network trained entirely in a simulated environment can generalize to flights in the real world.
In contrast with the previous methods based only on supervised learning, the authors additionally employ a reinforcement learning approach to derive a robust control policy from simulated data.
A 3D modeling suite is used to render various hallway configurations with randomized visual conditions.
Surprisingly, the control policy trained entirely in the simulated environment is able to navigate real-world corridors and rarely leads to collisions.

The work described above and other similar methods, for instance, \cite{giusti_machine_2016,gandhi_learning_2017,smolyanskiy_toward_2017}, use a data-driven approach to control a flying robot in real-world environments.
A shortcoming of these methods is that the learned controllers operate only in two-dimensional space which bears similar characteristics to navigation with ground robots.
Moreover, the approaches do not show the ability of the controllers to coordinate a multi-agent system.

\subsection{Vision-based multi-drone control}\label{sec:vision-flocking}

The control of multiple agents based on visual inputs is achieved with relative localization techniques \cite{saska_system_2017} for a group of three quadrotors.
Each agent is equipped with a camera and a circular marker that enables the detection of other agents and the estimation of relative distance.
The system relies only on local information obtained from the onboard cameras in near real-time.

Thus far, decentralized vision-based drone control has been realized by mounting visual markers on the drones \cite{saska_autonomous_2014,faigl_low-cost_2013,krajnik_practical_2014}.
Although this simplifies the relative localization problem significantly, the marker-based approach would not be desirable for real-world deployment of flying robots.
The used visual markers are relatively large and bulky which unnecessarily adds weight and drag to the platform; this is especially detrimental in real-world conditions.


\section{Method}\label{sec:method}

At the core of our method lies the prediction of a velocity command for each agent that matches the velocity command computed by a flocking algorithm as closely as possible.
For the remainder of the section, we consider the velocity command from the flocking algorithm as the target for a supervised learning problem.
The fundamental idea is to eliminate the dependence on the knowledge of the positions of other other agents by processing only local visual information.
Fig.~\ref{fig:flowchart} provides an overview of our method.

\subsection{Flocking algorithm}\label{sec:flocking-algorithm}

We use an adaptation of Reynolds flocking \cite{reynolds_flocks_1987} to generate targets for our learning algorithm.
In particular, we only consider the \tit{collision avoidance} and \tit{flock centering} terms from the original formulation since they only depend on relative positions.
We omit the \tit{velocity matching} term since estimating the velocities of other agents is an extremely difficult task given only a single snapshot in time (see Sec.~\ref{sec:dataset-generation}).
One would have to rely on estimating the orientation and heading with relatively high precision in order to infer velocities from a single image.

In our formulation of the flocking algorithm, we use the terms \tit{separation} and \tit{cohesion} to denote collision avoidance and flock centering, respectively \cite{saska_swarms_2014}.
We further add an optional \tit{migration} term that enables the agents to navigate towards a goal.

The first consideration when modeling the desired behavior of the flock is the notion of neighbor selection.
It is reasonable to assume that each agent can only perceive its neighbors in a limited range.
We therefore only consider agents as neighbors if they are closer than the desired cutoff distance $r^\txt{max}$ which corresponds to only selecting agents in a sphere with a given radius.
Therefore, we denote the set of neighbors of an agent $i$ as the set

\begin{equation}\label{eq:neighbor-set}
    \neighbors_i = \left\{ \txt{agents}~j : j \neq i \wedge \norm{\rel_{ij}} < r^\txt{max} \right\}
\end{equation}
where $\rel_{ij} \in \Real^3$ denotes the relative position of agent $j$ with respect to agent $i$ and $\norm{\cdot}$ the Euclidean norm.
We compute $\rel_{ij} = \norm{\pos_j - \pos_i}$ where $\pos_i \in \Real^3$ denotes the absolute position of agent $i$.

The separation term steers an agent away from its neighbors in order to avoid collisions.
The separation velocity command for the $i$th agent can thus be formalized as

\begin{equation}\label{eq:separation}
    \vel_i^\txt{sep} = -\frac{k^\txt{sep}}{|\neighbors_i|} \sum_{j \in \neighbors_i} \frac{\rel_{ij}}{\norm{\rel_{ij}}^2}
\end{equation}
where $k^\txt{sep}$ is the separation gain which modulates the strength of the separation between agents.

The cohesion term can be seen as the antagonistic inverse of the separation term since its purpose is to steer an agent towards its neighbors to provide cohesiveness to the group.
The cohesion velocity command for the $i$th agent can be written as

\begin{equation}\label{eq:cohesion}
    \vel_i^\txt{coh} = \frac{k^\txt{coh}}{|\neighbors_i|} \sum_{j \in \neighbors_i} \rel_{ij}
\end{equation}
where $k^\text{coh}$ is called the cohesion gain and modulates the tendency for the agents to be drawn towards the center of the neighboring agents.

For our implementation, the separation and cohesion terms are sufficient to generate a collision-free flock in which agents remain together, given that the separation and cohesion gains are chosen carefully.
We denote the combination of the two terms as the Reynolds velocity command $\vel_i^\txt{rey} = \vel_i^\txt{sep} + \vel_i^\txt{coh}$ which is later predicted by the neural network.

Moreover, the addition of the migration term provides the possibility to give a uniform navigation goal to all agents.
The corresponding migration velocity command is given by

\begin{equation}
    \vel_i^\txt{mig} = k^\txt{mig} \frac{\rel_i^\txt{mig}}{\norm{\rel_i^\txt{mig}}}
\end{equation}
where $k^\text{mig}$ denotes the migration gain and $\rel_i^\txt{mig} \in \Real^3$ denotes the relative position of the migration point with respect to agent $i$.
We compute $\rel_i^\txt{mig} = \pos^\txt{mig} - \pos_{i}$ where $\pos^\txt{mig} \in \Real^3$ is the absolute position of the migration point.

The velocity command for an agent $i$ is computed as a sum of the Reynolds terms, which is a combination of separation and cohesion, as well as the migration term, as $\tilde{\vel}_i = \vel_i^\txt{rey} + \vel_i^\txt{mig}$.
In general, we assume a homogeneous flock, which means that all agents are given the same gains for separation, cohesion, and migration.

A final parameter to adjust the behavior of the flock is the cutoff of the maximum speed.
The final velocity command that steers an agent is given by

\begin{equation}\label{eq:final}
    \vel_i = \frac{\tilde{\vel}_i}{\norm{\tilde{\vel}_i}} \min \left( \norm{\tilde{\vel}_i}, v^\txt{max}\right)
\end{equation}

where $v^{\txt{max}}$ denotes the desired maximum speed of an agent during flocking.

\subsection{Drone model}\label{sec:drone-model}

\begin{figure}[t]
\centering
\begin{subfigure}[b]{\columnwidth}
    \centering
    \includegraphics[width=0.96\textwidth]{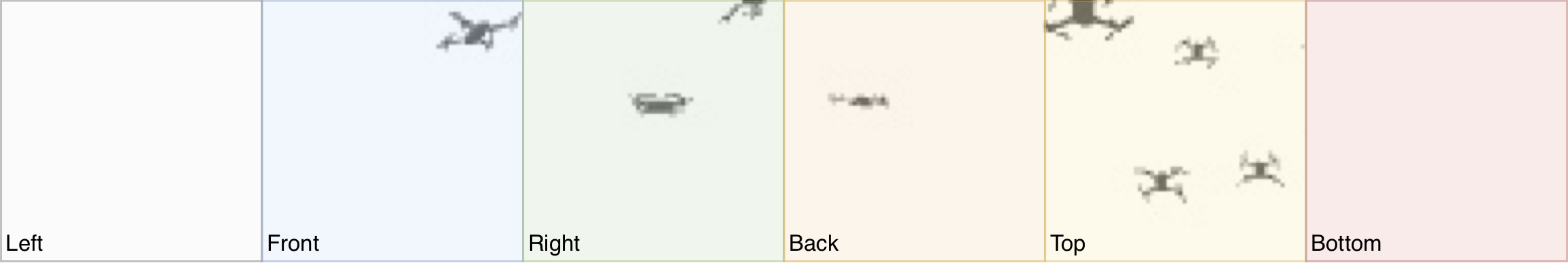}
    \caption{Visual input of an agent: concatenation of six camera images}\label{fig:field-of-view}
\end{subfigure}
\par\bigskip
\begin{subfigure}[b]{0.23\columnwidth}
    \centering
    \includegraphics[width=\textwidth]{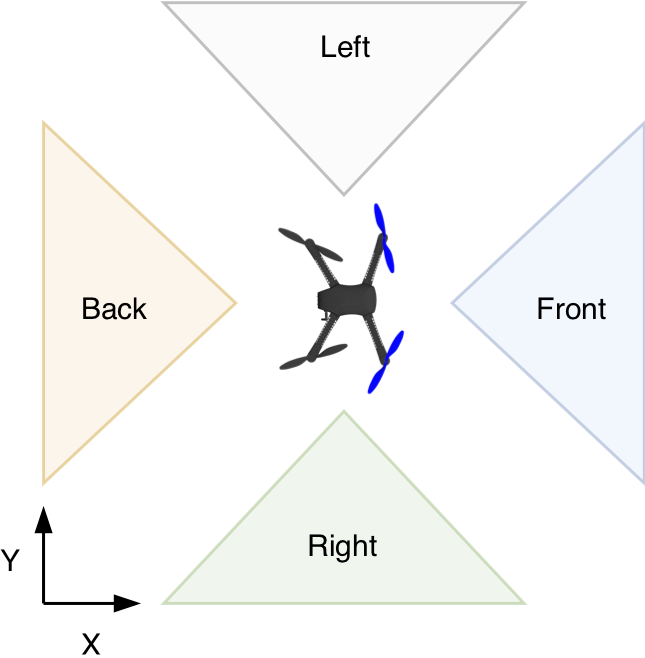}
    \caption{Top view}\label{fig:top-ortho}
\end{subfigure}
\quad
\begin{subfigure}[b]{0.23\columnwidth}
    \centering
    \includegraphics[width=\textwidth]{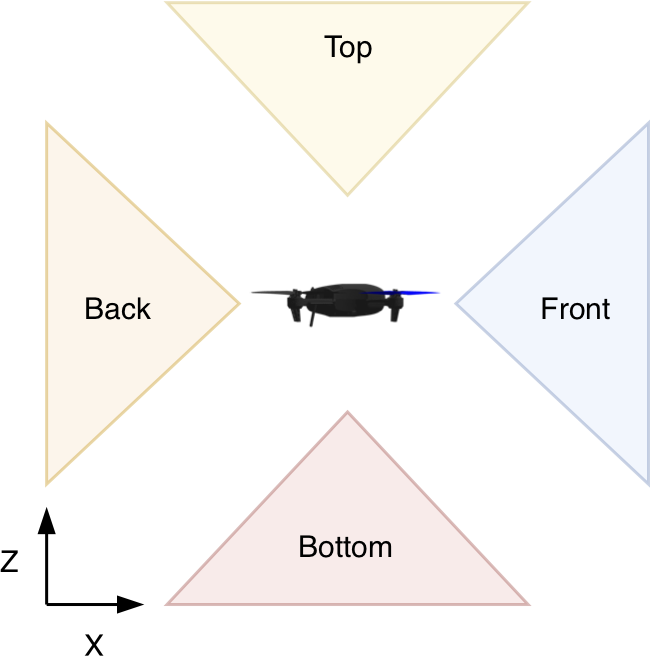}
    \caption{Side view}\label{fig:side-ortho}
\end{subfigure}
\quad
\begin{subfigure}[b]{0.42\columnwidth}
    \centering
    \includegraphics[width=\textwidth]{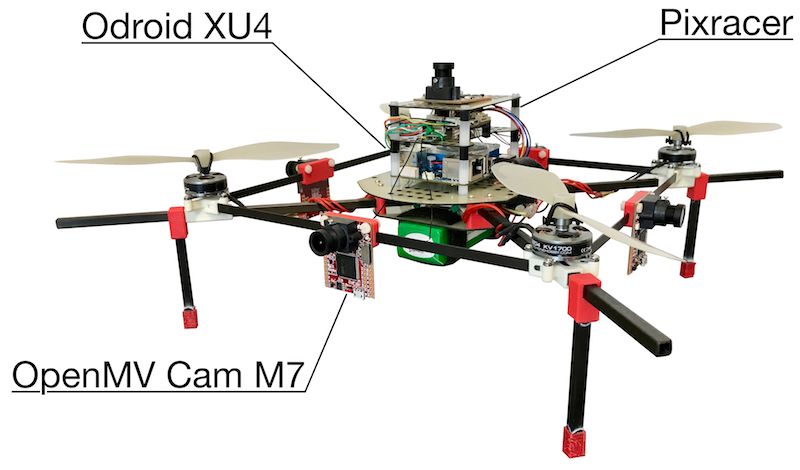}
    \caption{Drone hardware}\label{fig:drone-hardware}
\end{subfigure}
\caption{%
    Camera configuration and resulting visual input for a simulated agent.
    The cameras are positioned such that the visual field of an agent corresponds to a cube map, i.e. each camera is pointing at a different face of a cube as seen from within the cube itself.
    (\ref{fig:field-of-view}) The concatenation of the six camera images into the complete visual field.
    (\ref{fig:top-ortho}) and (\ref{fig:side-ortho}) Orthographic projection of the camera configuration as seen from the top and side, respectively.
    The schematic does not correspond to the exact geometry used in the simulation (see text for details).
    The color shades of the individual camera images (\ref{fig:field-of-view}) correspond to the field of view in the top (\ref{fig:top-ortho}) and side (\ref{fig:side-ortho}) orthographic projections.
    (\ref{fig:drone-hardware}) Example hardware implementation of the drone used in simulation based on \cite{dousse_human-comfortable_2017}.
    It uses six OpenMV Cam M7 for image acquisition, an Odroid XU4 for image processing, and a Pixracer autopilot for state estimation and control.
    }\label{fig:camera-config}
\end{figure}

Simulations were performed in Gazebo with a group of nine quadrotor drones.
Each drone is equipped with six simulated cameras which provide omnidirectional vision.
The cameras are positioned $15$\si{\cm} away from the center of gravity of the drone in order to have an unobstructed view of the surrounding environment, including the propellers (see Fig.~\ref{fig:field-of-view}).

Each camera has a $100^\circ$ horizontal and vertical field of view and takes a grayscale image of $64 \times 64$ pixels with a refresh rate of $10$ Hz.
We concatenate the images from all six cameras to a $64 \times 384 $ pixels grayscale image.

We use the PX4 autopilot \cite{meier_px4_2015} and provide it with velocity commands as raw setpoints via a ROS node. The heading of the drone is always aligned with the given velocity command.

\subsection{Dataset generation}\label{sec:dataset-generation}

We generate the dataset used for training the regressor entirely in a physics-based simulation environment.
Since our objective is to cover the maximum possible state and command space encountered during flocking, we generate our dataset with random trajectories as opposed to trajectories generated by the flocking algorithm itself.
In other words, we acquire an image $\Vector{x}_i$ and compute a ground truth velocity command $\target{\vel}_i$ from our flocking algorithm while following a random linear trajectory.
This was explicitly done to ensure that the dataset contains agent states which the flocking algorithm would not generate in order to improve robustness to unseen agent configurations.
Note that initial experiments with agents following trajectories generated by the flocking algorithm resulted in collisions between agents.
This observation is in line with the finding in \cite{ross_learning_2013} that situations not encountered during the training phase cannot possibly be learned by the controller.
A sample of the dataset is thus a tuple $\left( \Vector{x}_i, \target{\vel}_i \right)$ that is acquired for each agent $i$ every $10^{-1}$\si{\second}.
The dataset is generated in multiple \tit{runs}, each of which contains images and ground truth velocity commands generated by following a randomized linear trajectory as described below.
We generate $500$\si{\kilo} samples for training, $60$\si{\kilo} samples for validation, and $60$\si{\kilo} samples for testing.

The agents are spawned at random non-overlapping positions around the origin in a cube with side length of $4$\si{\meter} and a minimum distance to any other agent $1.5$\si{\meter}.
The side length and minimum distance were chosen to resemble a real-world deployment scenario of a drone swarm in a confined environment such as a narrow passage between adjacent buildings.
Each agent $i$ is then assigned a linear trajectory by following a velocity command which is drawn uniformly inside a unit cone with an angle of $15^\circ$.
The mean velocity command is thus facing directly in the direction of the migration point as seen from the origin.
The velocity command is distinct for each agent and kept constant during the entire run.
The random velocity commands were chosen such that collisions and dispersions are encouraged.

A run is considered complete as soon as a) the migration point is reached by at least one agent, or b) any pair of agents become either too close or c) too dispersed, all while following their linear trajectory.
We consider the migration point reached as soon as an agent comes within a radius $1$\si{\meter} of the migration point.
We consider the agents too close if any pair of drones falls below a collision threshold of $1$\si{\meter}.
Similarly, we regard agents as too dispersed when the distance between any two drones exceeds the dispersion threshold of $7$\si{\meter}.
The collision threshold follows the constraints of the drone model used in simulation and the dispersion threshold stems from the diminishing size of other agents in the field of view.

\subsection{Training phase}\label{sec:training-phase}

We formulate the imitation of the flocking algorithm as a regression problem which takes an image (see Fig.~\ref{fig:field-of-view}) as an input and predicts a velocity command which matches the ground truth velocity command as closely as possible.
To produce the desired velocities, we consider a state-of-the-art convolutional neural network \cite{loquercio_dronet_2018} that is used for drone navigation.
The model is composed of several convolutional layers with ReLU activations \cite{he_delving_2015} and finally a fully connected layer with dropout \cite{srivastava_dropout_2014} to avoid overfitting.
Unlike \cite{loquercio_dronet_2018} we opt for a single-head regression architecture to avoid convergence problems caused by different gradient magnitudes from an additional classification objective during training.
This simplifies the optimization problem and the model architecture and thus the resulting controller.

We use mini-batch stochastic gradient descent (SGD) to minimize the regularized mean squared error (MSE) loss between velocity predictions and targets as

\begin{equation}\label{eq:loss}
    \loss^\txt{MSE} = \frac{1}{\batchsize} \sum_{i = 1}^\batchsize (\target{\vel}_i - \pred{\vel}_i)^2 + \frac{\wdecay}{2} \norm{\Vector{w}}^2
\end{equation}
where $\target{\vel}_i$ is the target velocity and $\pred{\vel}_i$ the predicted velocity of the $i$th agent.
We denote the mini-batch size by $\batchsize$, the weight decay factor by $\wdecay$, and the neural network weights -- excluding the biases -- by $\Vector{w}$.
We employ variance-preserving parameter initialization by drawing the initial weights from a truncated normal distribution according to \cite{he_delving_2015}.
The biases of the model are initialized to zero.

The objective function is minimized using SGD with momentum $\momentum = 0.9$ \cite{sutskever_importance_2013} and an initial learning rate of $\lr = 5 \cdot 10^{-3}$ which is decayed by a factor of $k = 0.5$ after $10$ consecutive epochs without improvement on the hold-out validation set.
We train the network using a mini-batch size $\batchsize = 128$, a weight decay factor $\wdecay = 5 \cdot 10^{-4}$, and a dropout probability of $\pdropout = 0.5$.
We stop the training process as soon as the validation loss plateaus for more than ten consecutive epochs.

The raw images and velocity targets are pre-processed as follows.
Before feeding the images into the neural network, we employ global feature standardization such that the entire dataset has a mean of zero and a standard deviation of one.
For the velocity targets from the flocking algorithm, we perform a frame transformation from the world frame $\World$ into the drone's body frame $\Body$ as $\target{\vel_i} = {\Rotation^\Body_\World}_i \vel^\txt{rey}_i$ where ${\Rotation_\World^\Body}_i \in \SO{3}$ denotes the rotation matrix from world to body frame for robot $i$ and $\vel^\txt{rey}_i$ corresponds to the target velocity command.
We perform the inverse rotation to transform the predicted velocity commands from the neural network back into the world frame.

\subsection{Vision-based control}\label{sec:vision-based-control}

Once the convolutional network is finished training, we can use its predictions to modulate the velocity of the drone.
The same pre-processing steps apply to the vision-based control scenario, namely the standardization of raw images and the frame transformation of velocity predictions.
Although it is not mandatory for the implementation of this algorithm, one can optionally use a low-pass filter to smooth the velocity predictions, and as a result the trajectories of the agents.


\section{Results}\label{sec:results}

\newcommand{\cwidth}{0.04\textwidth}
\newcommand{\fwidth}{0.45\textwidth}

\begin{figure*}[t]
    \begin{subfigure}{\cwidth}
        \caption{}\label{fig:common-position}
    \end{subfigure}
    \begin{subfigure}{\fwidth}
        \includegraphics[width=\textwidth]{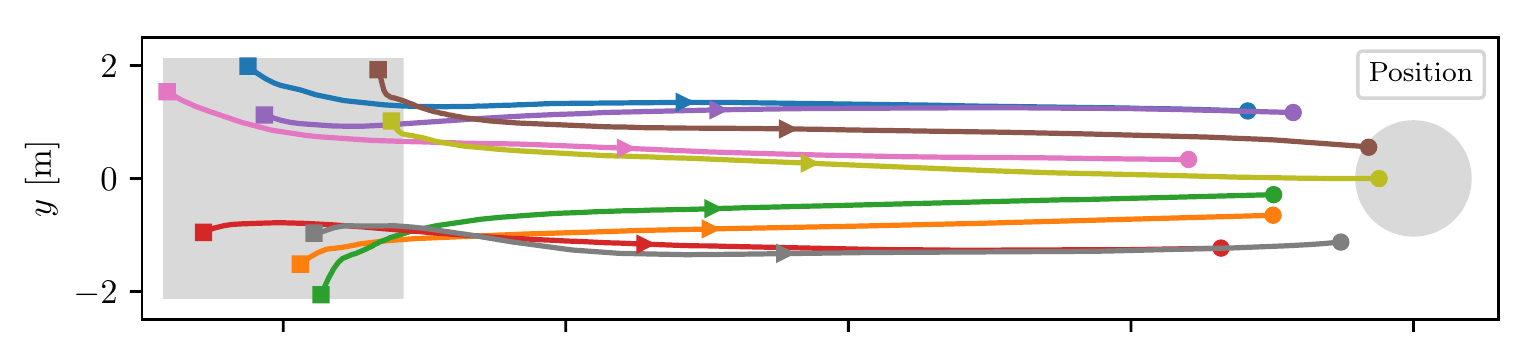}
    \end{subfigure}
    \hfill
    \begin{subfigure}{\fwidth}
        \includegraphics[width=\textwidth]{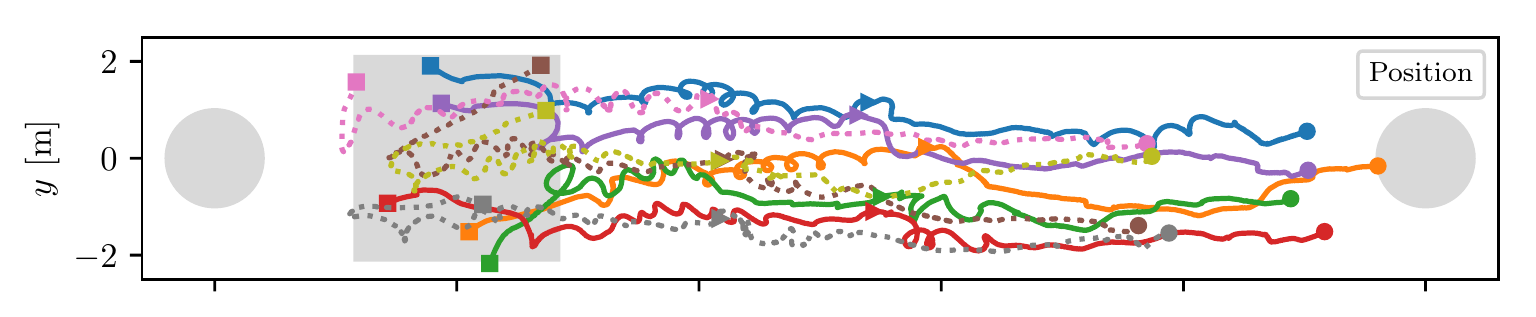}
    \end{subfigure}
    \begin{subfigure}{\cwidth}
        \caption{}\label{fig:opposing-position}
    \end{subfigure}
    \\
    \begin{subfigure}{\cwidth}
        \caption{}\label{fig:common-vision}
    \end{subfigure}
    \begin{subfigure}{\fwidth}
        \includegraphics[width=\textwidth]{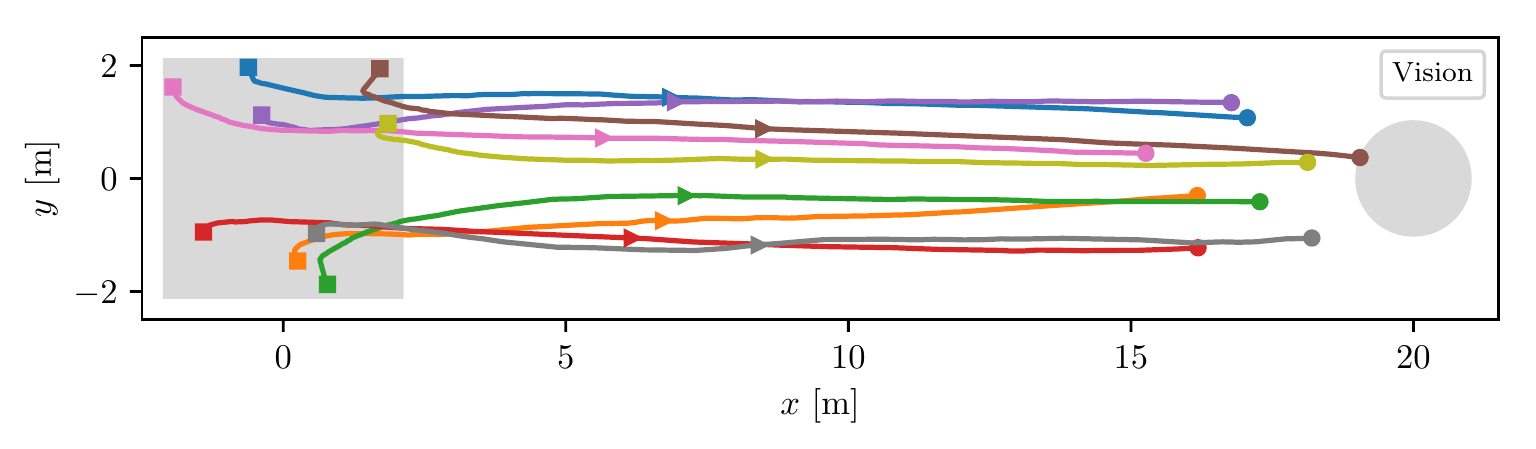}
    \end{subfigure}
    \hfill
    \begin{subfigure}{\fwidth}
        \includegraphics[width=\textwidth]{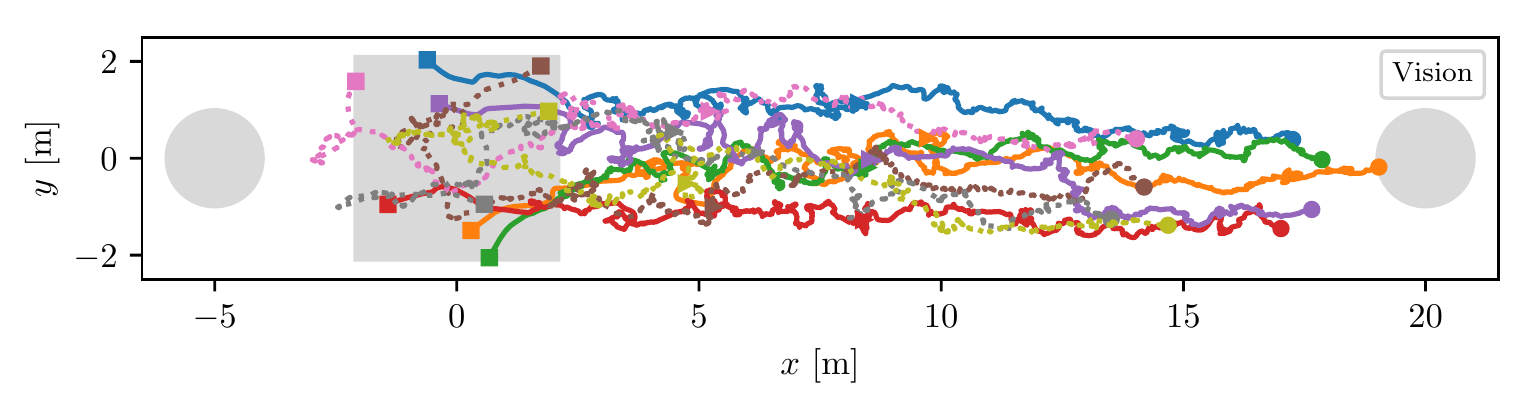}
    \end{subfigure}
    \begin{subfigure}{\cwidth}
        \caption{}\label{fig:opposing-vision}
    \end{subfigure}
    \\
    \begin{subfigure}{\cwidth}
        \caption{}\label{fig:common-distances}
    \end{subfigure}
    \begin{subfigure}{\fwidth}
        \includegraphics[width=\textwidth]{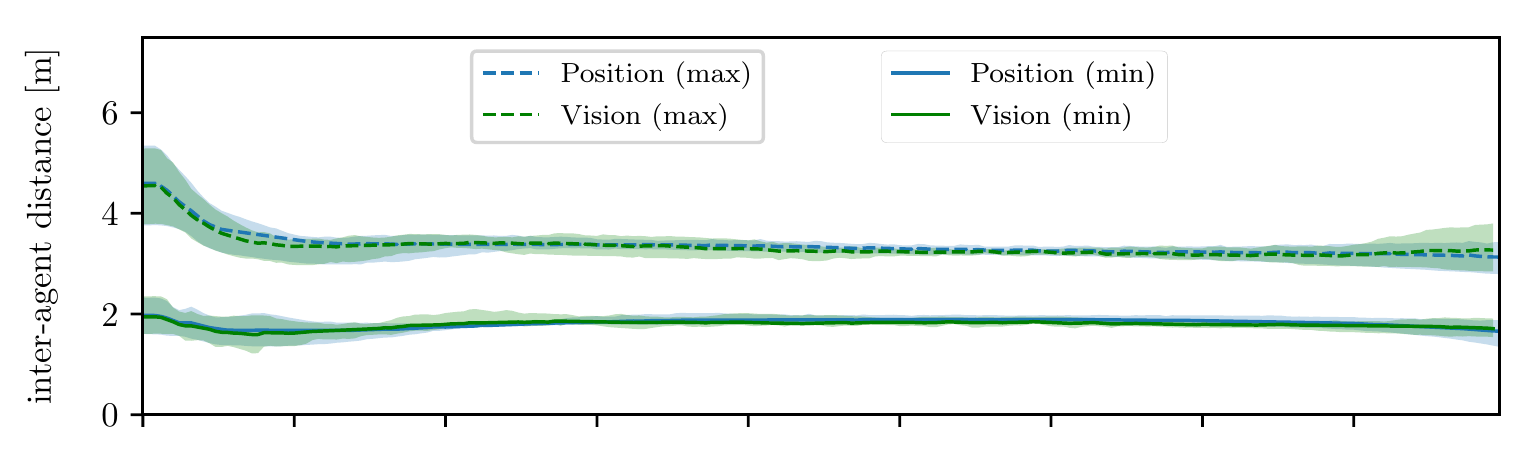}
    \end{subfigure}
    \hfill
    \begin{subfigure}{\fwidth}
        \includegraphics[width=\textwidth]{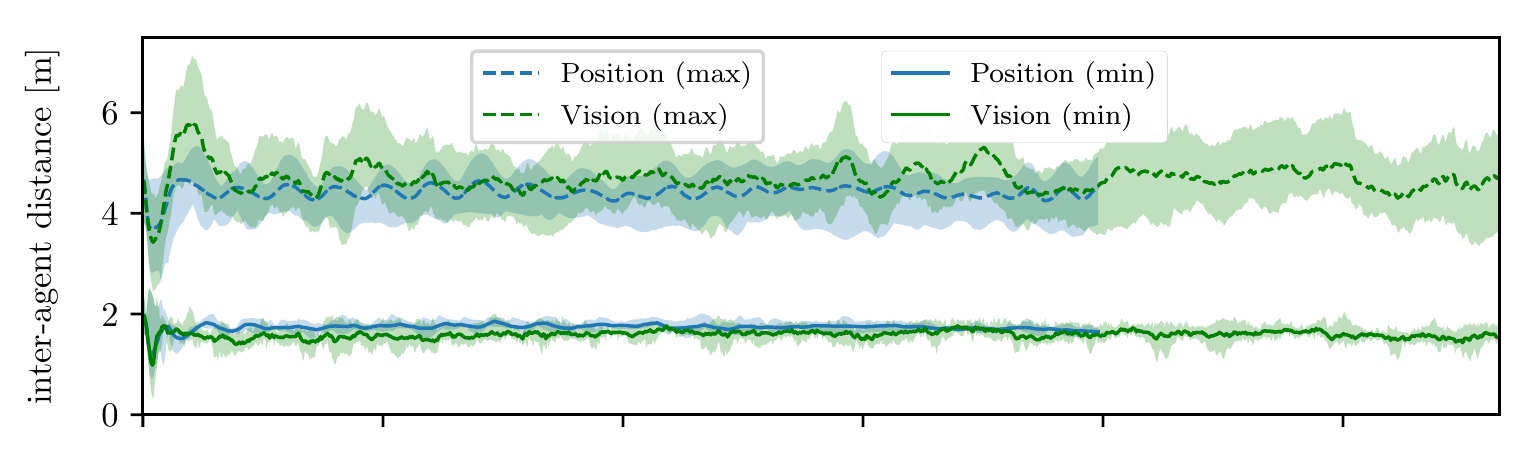}
    \end{subfigure}
    \begin{subfigure}{\cwidth}
        \caption{}\label{fig:opposing-distances}
    \end{subfigure}
    \\
    \begin{subfigure}{\cwidth}
        \caption{}\label{fig:common-order}
    \end{subfigure}
    \begin{subfigure}{\fwidth}
        \includegraphics[width=\textwidth]{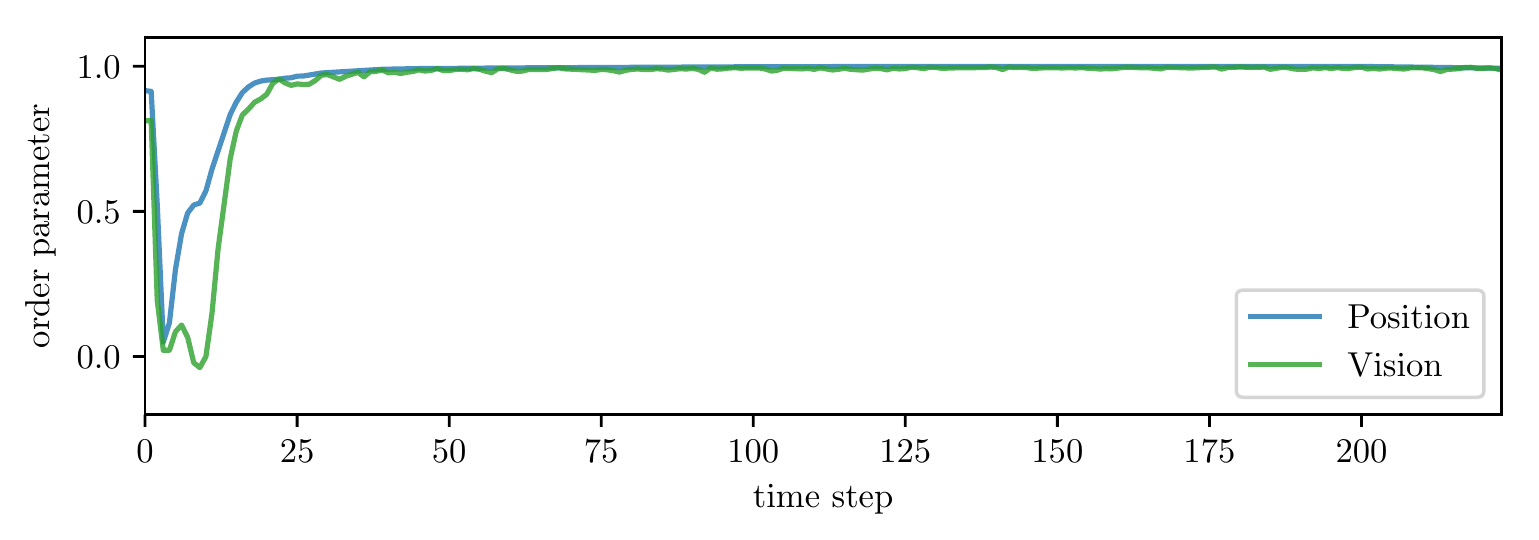}
    \end{subfigure}
    \hfill
    \begin{subfigure}{\fwidth}
        \includegraphics[width=\textwidth]{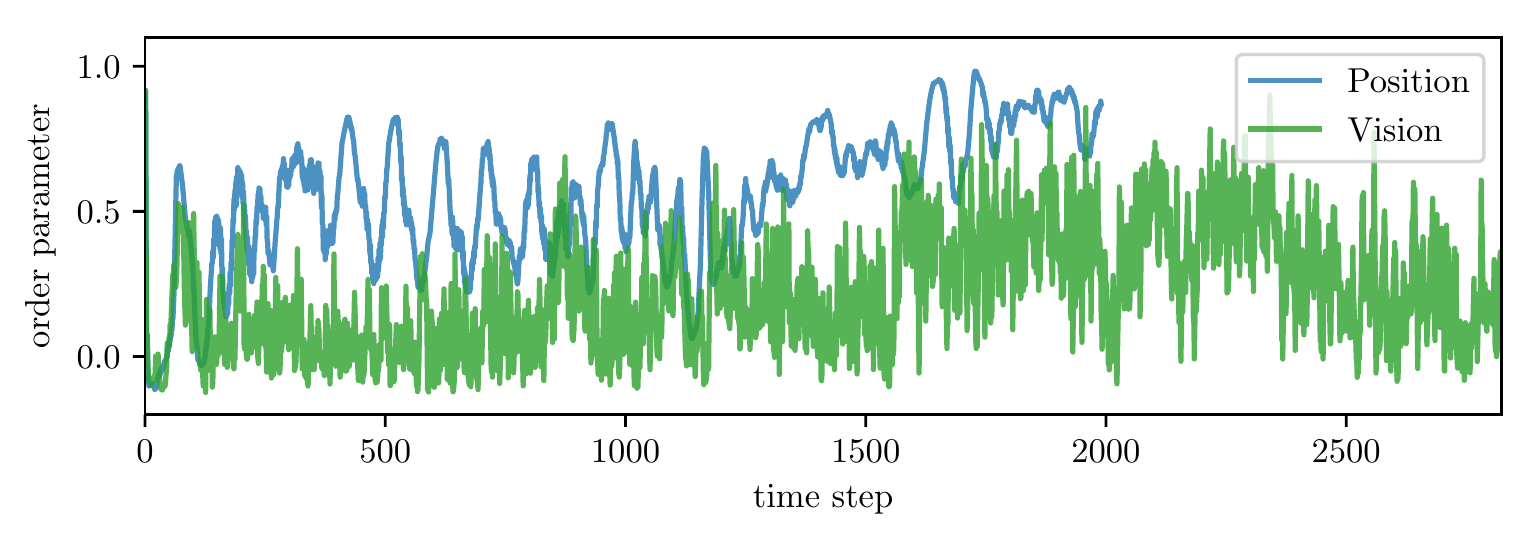}
    \end{subfigure}
    \begin{subfigure}{\cwidth}
        \caption{}\label{fig:opposing-order}
    \end{subfigure}
    \caption{%
        Flocking with common migration goal (left column) and opposing migration goals (right column).
        \textbf{First two rows}: Top view of a swarm migrating using the \tit{position-based} (\ref{fig:common-position} and \ref{fig:opposing-position}) and \tit{vision-based} (\ref{fig:common-vision} and \ref{fig:opposing-vision}) controller.
        The trajectory of each agent is shown in a different color.
        The colored squares, triangles, and circles show the agent configuration during the first, middle, and last time step, respectively.
        The gray square and gray circle denote the spawn area and the migration point, respectively.
        For the flock with opposing migration goal (\ref{fig:opposing-position} and \ref{fig:opposing-vision}), the waypoint on the right is given to a subset of five agents (solid lines), whereas the waypoint on the left is given to a subset of four agents (dotted lines).
        \textbf{Third row}: Inter-agent minimum and maximum distances from (\ref{eq:min-max-distance}) over time (\ref{fig:common-distances} and \ref{fig:opposing-distances}) while using the \tit{position-based} and \tit{vision-based} controller.
        The mean minimum distance between any pair of agents is denoted by a solid line, whereas mean maximum distances are shown as a dashed line.
        The colored shaded regions show the minimum and maximum distance between any pair of agents.
        \textbf{Fourth row}: Order parameter from (\ref{eq:order-parameter}) over time (\ref{fig:common-order} and \ref{fig:opposing-order}) during flocking.
        Note that the order parameter for the \tit{position-based} flock does not continue until the last time step since the \tit{vision-based} flock takes longer to reach the migration point.
    }\label{fig:migration-trajectories}
\end{figure*}
This section presents an evaluation of the learned vision-based swarm controller as a comparison to the behavior of the position-based flocking algorithm.
The results show that the proposed controller represents a robust alternative to communication-based systems in which the positions of other agents are shared with other members of the group.
We refer to the swarm operating on only visual inputs as \tit{vision-based} and the swarm operating on shared positions as \tit{position-based}.

The experiments are performed using the Gazebo simulator \cite{koenig_design_2004} in combination with the PX4 autopilot \cite{meier_px4_2015} for state estimation and control.
We employ the same set of flocking parameters used during the training phase throughout the following experiments (see Tab.~\ref{tab:flocking-params}).
Unless otherwise stated in the text, those parameters remain constant for the remainder of the experimental analysis.
The neural network is implemented in PyTorch \cite{paszke_automatic_2017}.
All simulations use the same random seed for repeatability.

\begin{table}
    \centering
    \caption{Parameters used for the flocking algorithm.}\label{tab:flocking-params}
    \begin{tabular}{@{}lllr@{}} \toprule
        Parameter       & Unit             & Description                & Value \\ \midrule
        $N$             & -                & Number of agents           & $9$   \\
        $r^\txt{max}$   & \si{\m}          & Maximum perception radius  & $20.0$\\
        $v^\txt{max}$   & \si{\m\per\s}    & Maximum flock speed        & $2.0$ \\
        $k^\txt{sep}$   & \si{\m\per\s}    & Separation gain            & $7.0$ \\
        $k^\txt{coh}$   & \si{\m\per\s}    & Cohesion gain              & $1.0$ \\
        $k^\txt{mig}$   & \si{\m\per\s}    & Migration gain             & $1.0$ \\ \bottomrule
    \end{tabular}
\end{table}

\subsection{Flocking metrics}\label{sec:flocking-metrics}

We report our results in terms of three flocking metrics that describe the state of the swarm at a given time step.
The  measures are best described as \tit{distance} and \tit{alignment}-based.

The two most important metrics are the minimum and maximum inter-agent \tit{distance} within the entire flock

\begin{equation}\label{eq:min-max-distance}
    d^\txt{min} = \min_{\substack{i, j \in \agents \\ i < j}} \norm{\rel_{ij}} \quad \txt{and} \quad
    d^\txt{max} = \max_{\substack{i, j \in \agents \\ i < j}} \norm{\rel_{ij}}
\end{equation}
where we let $\agents$ denote the set of all agents, and we have $i < j$ because of symmetry.
The minimum and maximum inter-agent distance are direct indicators for successful collision avoidance, as well as general segregation of the flock, respectively.
For instance, a collision occurs if the distance between any pair of agents falls below a threshold.
Similarly, we consider the flock too dispersed if the pairwise distance becomes too large.

We also measure the overall \tit{alignment} of the flock using an order parameter based on the cosine similarity

\begin{equation}\label{eq:order-parameter}
    \orderparam = \frac{1}{|\agents| (|\agents| - 1)} \sum_{i \in \agents} \sum_{\substack{j \in \agents \\ j \neq i}} \frac{{\vel_i}^\top \vel_j}{\norm{\vel_i} \norm{\vel_j}}
\end{equation}
which measures the degree to which the heading of the agents agree with each other \cite{viragh_flocking_2014}.
If the headings are aligned, we have $\orderparam \approx 1$ and in a disordered state, we have $\orderparam \approx 0$.
Recall that the agent's body frame is always aligned with the direction of motion.

\subsection{Common migration goal}\label{sec:common-migration-goal}

In the first experiment, we give all agents the same migration goal and show that the swarm remains collision-free during navigation.
Both the \tit{vision-based} and the \tit{position-based} swarm exhibit remarkably similar behavior while migrating (see Figs.~\ref{fig:common-position} and \ref{fig:common-vision}).
For the \tit{vision-based} controller, one should notice that the velocity commands predicted by the neural network are sent to the agents in their raw form without any further processing.

We are especially interested in the minimum and maximum inter-agent distances, i.e. the extreme distances between any pair of agents, during migration.
The minimum distance can be used as a direct measure for collision avoidance, whereas the maximum distance is a helpful metric when deciding whether or not a swarm is coherent.
The \tit{vision-based} controller matches the \tit{position-based} one very well since they do not deviate significantly over the course of the entire trajectory (see Fig.~\ref{fig:common-distances}).
If the neural controller had not learned to keep a minimum inter-agent distance, we would observe collisions in this case.

\subsection{Opposing migration goals}\label{sec:opposing-migration-goals}

In this experiment, we assign different migration goals to two subsets of agents.
The first group, consisting of five agents, is assigned the same waypoint as in Sec.~\ref{sec:common-migration-goal}.
The second group, consisting of the remaining four agents, is assigned a migration point on the opposite side with respect to the first group.
The \tit{position-based} and \tit{vision-based} flock exhibit very similar migration behaviors (see Figs.~\ref{fig:opposing-position} and \ref{fig:opposing-vision}).
In both cases, the swarm cohesion is strong enough to keep the agents together despite the diverging navigational preferences.
This is the first sign that the neural network learns the non-trivial difference between agents that are too close or too far away.

We can observe slightly different behaviors between the two modalities when measuring the alignment within the flock.
The \tit{position-based} flock tends to be ordered to a greater extent than its \tit{vision-based} counterpart (see Fig.~\ref{fig:opposing-order}).
This periodicity in the order of the flock stems from circular motion exhibited by the \tit{position-based} agents (see Fig.~\ref{fig:opposing-position}).
There is less regularity in the \tit{vision-based} flock, in which agents tend to be less predictable in their trajectories, albeit remaining collision-free and coherent.
Note that the \tit{vision-based} flock tends to be less well aligned and also reaches its migration goal far later than the \tit{position-based} flock.

\subsection{Generalization of the neural controller}\label{sec:generalization}

We performed a series of ablation studies to show that the learned controller generalizes to previously unseen scenarios.
These experiments can be seen as perturbations to the conditions that the agents were exposed to during the training phase.
This allows us to highlight failure cases of the controller as well as show its robustness towards changing conditions.
We change parameters such as the number of agents $N = \{3, 12\}$ in the swarm or the maximum flock speed $v^\mtx{max} = \{ 2.0, 4.0\}$\si{\meter\per\second}.
We also remove the external migration point $\pos^\mtx{mig}$ and the corresponding term entirely to show how the flock behaves when it is self-propelled.

The swarm remains collision-free and coherent without migration point or when the number of agents and the flock speed is changed.
We could not observe a noticeable difference in the inter-agent distance or the order parameter during the experiments.

\subsection{Attribution study}\label{sec:attribution-study}

\begin{figure}
    \centering
    \includegraphics[width=\columnwidth]{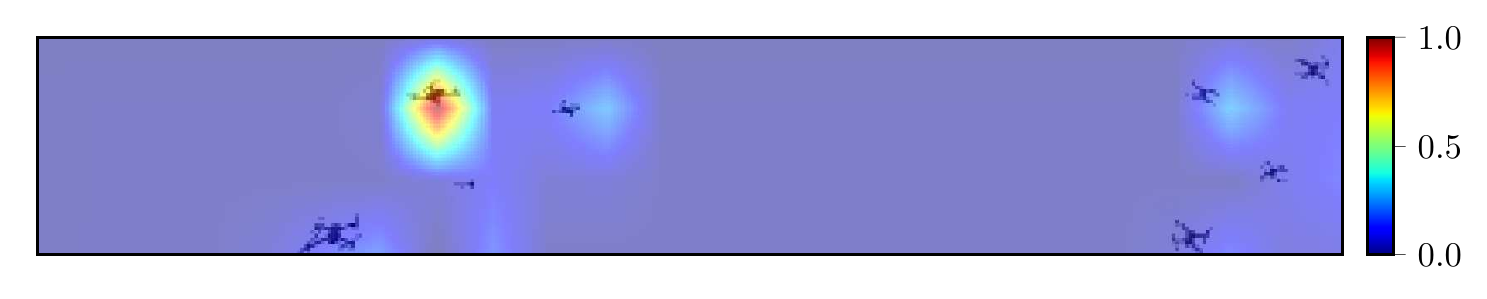}
    \caption{%
        Heat map visualization of the relative importance of each pixel in the visual input of an agent towards the velocity command.
        Red regions have the most influence on the control command, whereas blue regions contribute the least.
    }\label{fig:heatmap}
\end{figure}

Since the \tit{vision-based} controller provides a very tight coupling between perception and control, the need for interpretation of the learned behavior arises.
To this end, we employ a state-of-the-art attribution method \cite{selvaraju_grad-cam_2017}, which shows how much influence each pixel in the input image has on the predicted velocity command (see Fig.~\ref{fig:heatmap}).
More specifically, we compute the gradients for the heat map by backpropagating with respect to the penultimate strided convolutional layer of the neural network in which the individual feature maps retain a spatial size of $4 \times 24$ pixels.
We then employ bilinear upsampling to increase the resolution of the resulting saliency map before we blend it with the original input image.
We attribute the relatively poor localization performance of some of the agents to the low spatial acuity of the generated heat map.

We can observe a non-uniform distribution of importance that seems to concentrate on a single agent that is located in the field of view of the front-facing camera (see Fig.~\ref{fig:field-of-view}).
The image is taken from an early time step during migration where the magnitudes of the predicted velocity commands are relatively large.
We notice that the network is effectively localizing the other agents spatially in the visual input, albeit having not been explicitly given the positions as targets.
The saliency map is generated very efficiently by computing the backward pass until the target convolutional layer and could therefore serve as a valuable input to a real-time detection algorithm.

\section{Conclusions and Future Work}\label{sec:conclusions}

This paper presented a machine learning approach to the problem of collision-free and coherent motion of a dense swarm of quadcopters.
The agents learn to coordinate themselves \tit{only} via visual inputs in 3D space by mimicking a flocking algorithm.
The learned controller removes the need for communication of positions among agents and thus presents the first step towards a fully decentralized vision-based swarm of drones.
The trajectories of the flock are relatively smooth even though the controller is based on \tit{raw} neural network predictions.
We show that our method is robust to perturbations such as changes in the number of agents the or maximum speed of the flock.
Our algorithm naturally handles navigation tasks by adding a migration term to the predicted velocity of the neural controller.

We are actively working on transferring the flock of quadrotors from a simulated environment into the real world.
A motion capture system is used to generate the dataset of ground truth positions and real camera images, similar to the simulation setup described in this paper.
A natural subsequent step will be the transfer of the learned controller to outdoor scenarios where ground truth positions will be obtained using a GNSS.
To reduce the need for large amounts of labeled data, we are exploring recent advances in deep domain adaptation \cite{ganin_domain-adversarial_2016,rozantsev_beyond_2018} to aid generalization of the neural controller to environments with background clutter.
Another challenge is the addition of obstacles to the environment in which the agents operate.
To this end, we will opt for more sophisticated flocking algorithms which allow the direct modeling of obstacles \cite{viragh_flocking_2014,vasarhelyi_optimized_2018}, as well as pre-defining the desired distance between agents \cite{olfati-saber_flocking_2006}.


\section{Acknowledgements}\label{sec:acknowledgements}

\ifblind
    Intentionally left blank for double-blind review purposes.
\else
    \blind{We thank Enrica Soria for the feedback and helpful discussions, as well as Olexandr Gudozhnik and Przemyslaw Kornatowski for their contributions to the drone hardware.}
    \blind{This research was supported by the Swiss National Science Foundation (SNF) with grant number 200021\_155907 and the Swiss National Center of Competence Research (NCCR).}
\fi

\fontsize{9.0pt}{10.0pt}\selectfont
\bibliography{library/strings/conferences-abrv,library/strings/journals-abrv,library/library}
\bibliographystyle{aaai}

\end{document}